\documentclass[runningheads]{llncs}
\usepackage{color}
\usepackage[T1]{fontenc}
\usepackage{graphicx}
\usepackage{multirow}
\usepackage{amsmath}
\begin{document}
\title{Complex Ontology Matching with Large Language Model Embeddings}
%


\author{Guilherme Santos Sousa\inst{1}\orcidID{0000-0002-2896-2362} \and
Rinaldo Lima\inst{2}\orcidID{0000-0002-1388-4824} \and Cassia Trojahn\inst{1}\orcidID{0000-0003-2840-005X}}
\authorrunning{Sousa et al.}
%
\institute{IRIT \& Universit\'e de Toulouse 2 Jean Jaur\`es, Toulouse, France \and
Universidade Federal Rural de Recife, Recife, Brazil\\
}
\maketitle              
\begin{abstract}
Ontology, and more broadly, Knowledge Graph Matching is a challenging task in which expressiveness has not been fully addressed. Despite the increasing use of embeddings and language models for this task, approaches for generating expressive correspondences still do not take full advantage of these models, in particular, large language models (LLMs). This paper proposes to integrate LLMs into an approach for generating expressive correspondences based on alignment need and ABox-based relation discovery. The generation of correspondences is performed by matching similar surroundings of instance sub-graphs. The integration of LLMs results in different architectural modifications, including label similarity, sub-graph matching, and entity matching. The performance word embeddings, sentence embeddings, and LLM-based embeddings, was compared. The results demonstrate that integrating LLMs surpasses all other models, enhancing the baseline version of the approach with a 45\% increase in F-measure.

\end{abstract}
\section{Introduction}

Ontology matching (and more broadly, knowledge graph matching) aims at enabling interoperability between knowledge expressed in different schemes. This task is at the core of knowledge graph-oriented applications. While the ontology matching field has reached some maturity, most of the matching approaches still focus on generating simple correspondences (i.e., those linking one single entity of a source ontology to one single entity of a target ontology, as $Authors$ $\equiv$ $Writer$). However, this type of correspondence is not expressive enough to fully cover the different kinds of heterogeneities (lexical, semantic, conceptual, granularity) from different schemes. The need for complex correspondences (i.e., those involving logical constructors or transformation functions, as e.g., $Accepted\_Paper$ $\equiv$ $Paper$ $\sqcap$ $\exists$ hasDecision.Acceptance) has been recognized across various fields, such as cultural heritage \cite{nurmikko-fuller_building_2015}, agronomic \cite{thieblin2017cross}, or still biomedical \cite{jouhet2017building}.

With the rise of language models, recent matching approaches rely on such models \cite{DBLP:conf/semweb/SousaLT22}. This increased adoption is due to their capability of modeling the textual and structural information present in ontologies and knowledge graphs. These representations in the form of embeddings can produce better similarity scores.
However, these models have not been fully explored for complex matching. The main issues involve defining entity boundaries, the scarcity of datasets with reference alignments for training supervised models, or still finding good representation for complex entities like unions or restrictions. More recently, Large Language Models (LLMs) have achieved higher levels of contextual and semantic understanding of the text and are capable of producing better representations for similarity tasks \cite{DBLP:conf/eacl/MuennighoffTMR23,DBLP:journals/corr/abs-2307-03393,DBLP:journals/corr/abs-2404-10329}.

This paper takes advantage of the latest advancements in LLMs for the task of complex matching. It extends CANARD \cite{thieblincanard2024}, an approach that relies on knowledge graphs (entities) equipped with ontologies (schema), and a search space reduction strategy guided by user knowledge needs (expressed as SPARQL queries) in terms of alignment. It takes as input a SPARQL query over the source ontology and matches the subgraph from the source SPARQL to the lexically similar surroundings of the instances from the target ontology. Four architectural modifications are proposed: label embedding similarity, embeddings of SPARQL results, subgraph embeddings, and instance embeddings. They address the issues highlighted earlier, and pre-trained models have been adopted to circumvent the necessity for reference alignments. Moreover, refining the graph search based on user knowledge aids in better defining entity boundaries and representation aggregations to describe complex constructors. Experimental results demonstrate that the proposal significantly enhances not only the performance compared to the baseline approach but also when contrasted with state-of-the-art ones.

The contributions of this paper can be summarised as follows: (i) an entirely revised approach for complex matching that takes advantage of the strengths of LLM embeddings in different key steps of the matching process; (ii) a comprehensive analysis of the impact of using such embeddings in each architectural modification, on well-known benchmarks in complex matching; (iii) a comparison of the proposed approach to state-of-the-art ones; and (iv) a discussion of their strengths and weaknesses.

The rest of this paper is structured as follows. Section \ref{sec:approach_overview} presents the proposed approach. Section \ref{sec:experiments} describes the experimental setting. Section \ref{sec:results} presents the results and discusses them. Related work is discussed in Section \ref{sec:related_work}. Finally, Section \ref{sec:conclusions} concludes the paper and outlines paths to future directions.

\section{Proposed Approach}
\label{sec:approach_overview}

\subsubsection{Baseline approach} 
\begin{figure}
    \centering
    \includegraphics[width=\textwidth]{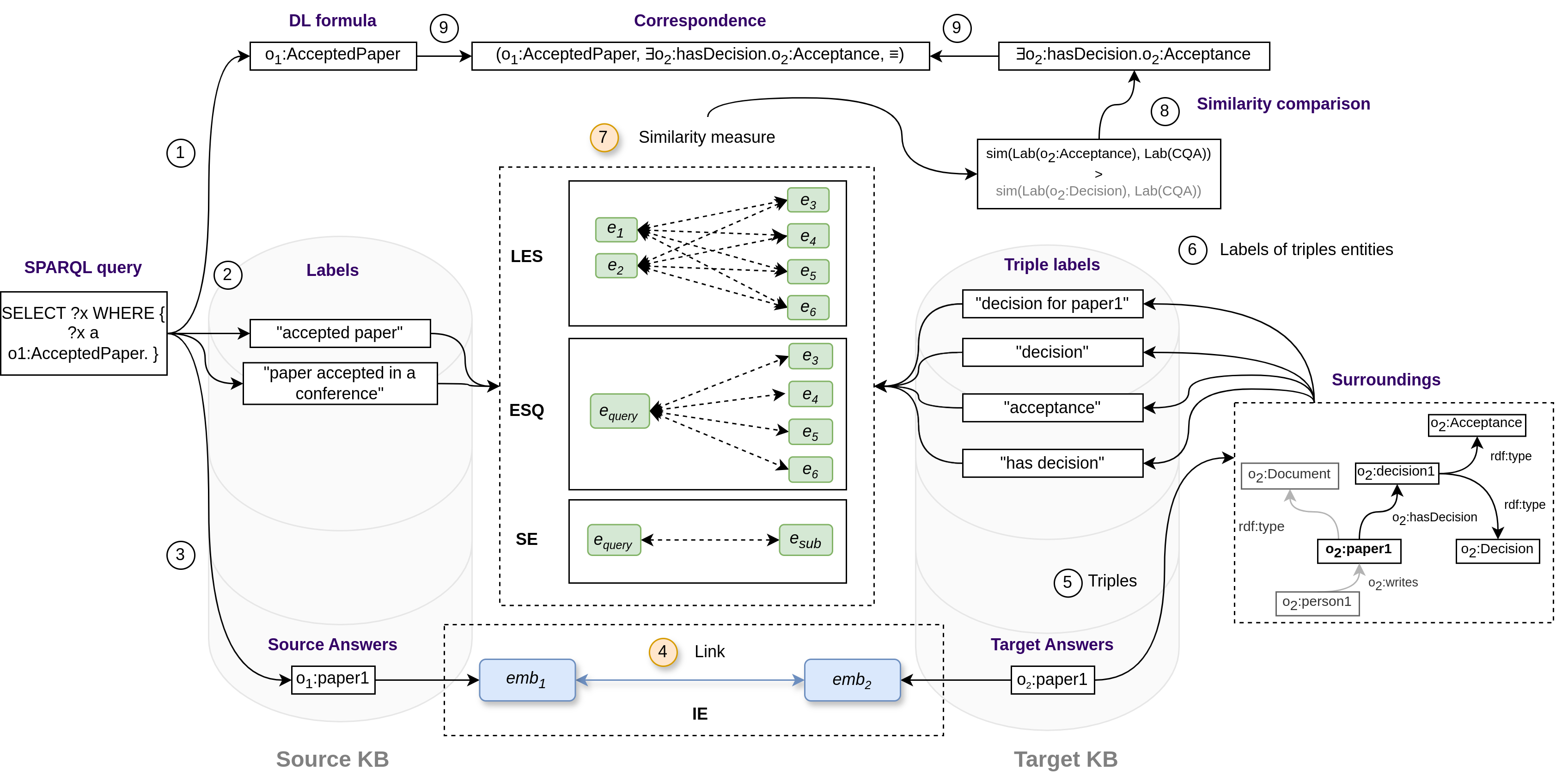}
    \caption{Original architecture and steps where embeddings have been used. In the Figure, LES refers to Label embedding similarity, ESQ to Embeddings of SPARQL query, SE to Subgraph embeddings, and IE Instance embeddings.}
    \label{fig:narch}
    \vspace{-0.5cm}
\end{figure}

CANARD takes as input a set of SPARQL SELECT queries over the source ontology, which express the user needs in terms of alignment. The reader can refer to \cite{thieblincanard2024} for details. According to query arity, three types of queries are considered: a \textit{unary} question expects a set of instances, e.g., ``Which are the accepted papers?'' \textit{(paper1), (paper2)}; a \textit{binary} question expects a set of instances or value pairs, e.g., ``What is the decision on a paper?'' \textit{(paper1, accept), (paper2, reject)}; and a\textit{n-ary} question expects a tuple of size $\geq$ 3, e.g., ``What is the decision associated with the review of a given paper?'' \textit{(paper1, review1, weak accept), (paper1, review2, reject)}. Queries for the approach are limited to \textit{unary} and \textit{binary} questions, of \textit{select} type, and no modifier\footnote{This is a limitation in the sense that we do not deal with specific kinds of SPARQL queries, as the ones involving CONSTRUCT and ASK. The approach does not deal with transformation functions or filters inside the SPARQL queries and only accepts queries with one or two variables. However, as classes and properties are unary and binary predicates, these limitations still allow the approach to cover ontology expressiveness.}. CANARD requires that the source and target ontologies have an $\mathcal{A}box$ with at least one common instance for each SPARQL query. The overall approach is articulated in 9 steps (Figure \ref{fig:narch}). These steps and the subsequent modifications are introduced in the rest of the paper. Overall, the matching is performed by finding the surroundings of the target instances which are lexically similar to the CQA. The hypothesis behind the approach is to rely on a few examples (answers) to find a generic rule that describes more instances. 

\subsubsection{Revised approach} 

In the baseline approach, two core steps require the lexical comparison of entity labels: (1) finding common instances between source and target KG {(Step 4 in Figure \ref{fig:narch})}; and (2) computing the similarity between SPARQL query labels and the target subgraphs retrieved from the common instances found in the linking step {(Step 7 in Figure \ref{fig:narch})}. To enhance the effectiveness of similarity methods in these steps, the proposal is to incorporate embedding similarity. LLMs are employed to encode textual information and generate embeddings for each node in the KG. The hypothesis is that leveraging LLMs can enhance the overall matching performance, as LLMs have been demonstrated to yield superior embedding features, particularly in KG applications. The proposed approach is available in GitLab \footnote{https://gitlab.irit.fr/melodi/ontology-matching/complex/llm-embedding-complex}.

\subsubsection{Embedding Generation}

\begin{figure}[!ht]
    \centering
    \includegraphics[width=\textwidth]{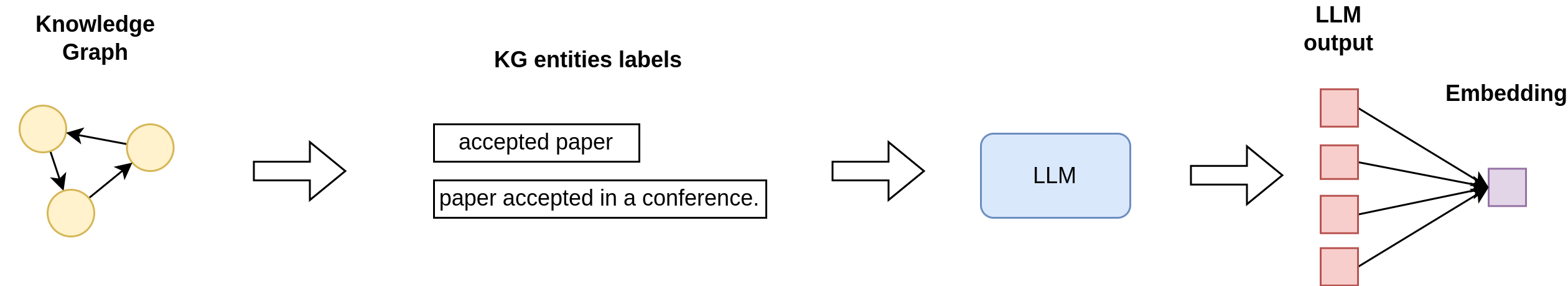}
    \caption{Process of generating embeddings from a given label. Each label is tokenized, then the tokens are fed into the language model to generate the embeddings for each token. The last step is aggregating all token embeddings to generate a label embedding.}
    \label{fig}
    \vspace{-0.5cm}
\end{figure}

The embeddings are prepared before the matching process starts. The ontology is loaded, and all labels from the ontology entities are processed to generate the embeddings, as illustrated in Figure \ref{fig}. First, each label is tokenized into individual tokens using the model default tokenizer. {Then the tokens are fed into the LLM that produces embeddings as output. The last hidden layer of this output is averaged to produce the final embedding of the label.} Specifically, the output from the last hidden layer is a tensor of dimensions (B, S, N), where B is the batch size, S is the number of tokens in the label, and N is the embedding dimension. Second, to generate a single embedding vector for each label, the embeddings are averaged over the sentence dimension (S). This results in a final embedding of dimensions (B, 1, N), producing a fixed-size embedding for each label. This process is uniform across all models, regardless of their architecture (encoder-only, decoder-only, or encoder-decoder), as all models provide a last hidden layer output that can be averaged. 

\subsubsection{Matching Step}

The proposed pipeline is presented in Figure \ref{fig:narch} where one of the major differences from the baseline approach concerns the three variations of how the embeddings are combined in the subgraph similarity step (step 7 in Figure \ref{fig:narch}). The overall approach leverages a SPARQL query to guide the alignment of entities within the query context, effectively reducing the matching space. This process involves using the query to retrieve instances that serve as anchors for identifying common subgraphs between the source and target datasets. These subgraphs, which can consist of triples or paths (complex entities), vary depending on the query type (unary or binary). Unary queries result in triple subgraphs, while binary queries yield path subgraphs. Consequently, different aggregation methods are applied to each subgraph type (Figure \ref{fig:triplex}). Subsequently, the identified subgraphs are compared based on their similarity to the entities mentioned in the query text. For instance, a query regarding accepted papers might include the class 'Paper', since accepted papers are instances of this class. Alongside the class 'Paper', descriptions are retrieved by querying predicates such as \textit{skos\#prefLabel} or \textit{skos\#literalForm} associated with this class.

Embeddings are applied at various stages of the approach. The Instance Embeddings (IE) step occurs during the identification of common instances to retrieve subgraphs, managing the embeddings of these instances. In the step of comparing the similarity between the entities in the query and the retrieved subgraphs multiple potential similarity combinations (N elements in the query by M elements in the subgraph), arise and three strategies for aggregating embeddings are compared: The Label Embedding Similarity (LES), Embeddings of SPARQL Query (ESQ), and Subgraph Embeddings (SE). Detailed descriptions of each setting are provided below.

\subsection{Instance embeddings (IE)}

The Instance Embeddings (IE) setting involves integrating embeddings into the process of identifying common instances between the different KGs. In the baseline approach, the SPARQL query is consulted in the source KG and the resulting instances are used to find the corresponding instances in the target KG. These instances will serve as anchor links between the source and target knowledge graphs and used to retrieve the subgraphs in target KG. {For this search, the baseline architecture seeks for common instances associated by predicates such as rdf-schema\#seeAlso, owl\#sameAs, skos/core\#closeMatch, or skos/core\#exactMatch}. If these predicates are not found, an exact string matching is performed between the source and target KG {(step 4 in Figure \ref{fig:narch})}.

Given the unreliability of exact string matching in finding similar data within this context, incorporating embeddings can enhance the process of identifying similar instances between the knowledge bases. In the proposed approach, embeddings for the resulting entities are retrieved and stacked. Then to find similar instances, the labels of source query instances are embedded, and a cross-cosine similarity is computed between the instance embeddings in the source and the stacked embeddings from the target KG, resulting in a similarity matrix. Then, the instance with the highest similarity score in this matrix is selected, and, if it surpasses the link similarity threshold, it is returned as a link between the instance in source KG and the corresponding instance found in the target KG.

\subsection{Similarity step settings}

\begin{figure}[!ht]
    \centering
    \includegraphics[width=\textwidth]{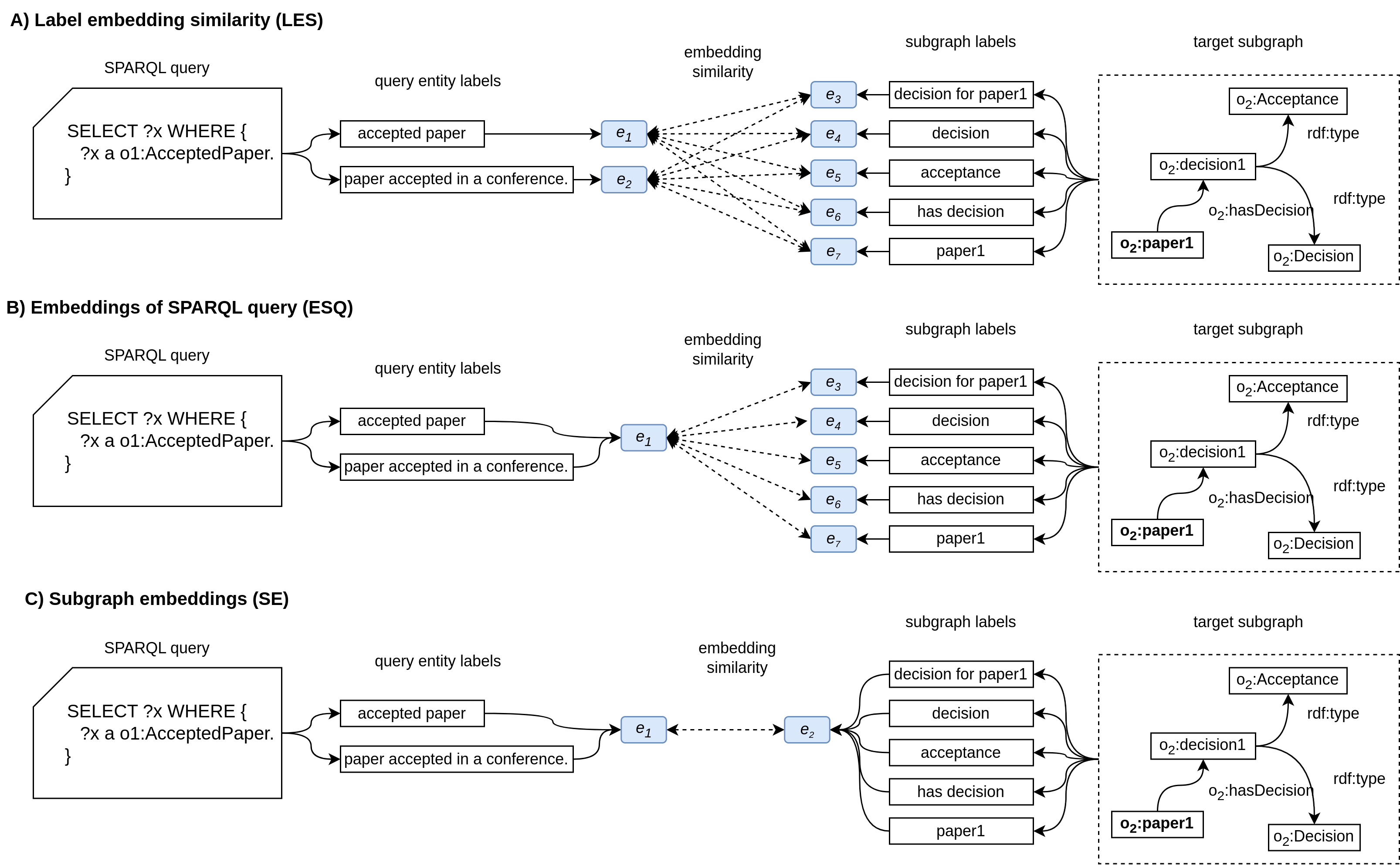}
    \caption{Example of the first three modifications in similarity computation. \textbf{Label embedding similarity} is the first modification where the Levenshtein similarity is replaced by embeddings similarity resulting in an n:m comparison. \textbf{Embeddings of SPARQL query} first the SPARQL query embeddings are aggregated and then compared with embeddings of subgraph labels resulting in a 1:m comparison. \textbf{Subgraph embeddings} are created by aggregating the embeddings of labels, resulting in a 1:1 comparison.}
    \label{fig:embm}
\end{figure}

In this step, the Levenshtein similarity metric is replaced with embedding similarity (step 7 in Figure \ref{fig:narch}), which can be configured in three different settings as depicted in Figure \ref{fig:embm}. These settings alter the level of aggregation of the embeddings before comparison and are progressively applied while keeping each modification from the previous ones. A threshold filter is applied to determine which parts of the subgraphs contribute to the final similarity value, allowing for similarity values greater than 1 in the final value. 

\subsubsection{Label embedding similarity (LES)}
\label{sec:label-emb-sim}

Previously, in this step, all labels retrieved from the entities present in the SPARQL query source {SELECT distinct ?s WHERE \{ ?s a <:AcceptedPaper>. \}} (e.g., "accepted paper", "paper accepted in a conference") are compared with the labels of the entities in the subgraphs (Step 7 in Figure \ref{fig:narch}) retrieved after the linking step (Step 4 in Figure \ref{fig:narch}) using the Levenshtein similarity. This resulted in $N \times M$ computations, where $N$ denotes the labels associated with the SPARQL query, and $M$ represents the labels from the entities in the target subgraphs. The final similarity is calculated as the sum of all similarity values between the labels in the Cartesian product $N \times M$, with similarities below a certain threshold being filtered out. The Levenshtein metric is known to introduce false positive correspondences that are lexically similar but not equivalent (e.g., "Review" and "Reviewer"). By replacing this metric with similarity derived from embedding representations, a better comparison of semantic information in labels becomes feasible, as false positives can be filtered out by assessing the semantic meaning present in the embeddings. In the new architecture, embeddings are employed to compute the similarity between the labels associated with the SPARQL queries and the labels in target subgraphs. An embedding is retrieved for each label before comparison and then the similarity between the embeddings of labels is calculated using cosine similarity. The same steps of threshold filtering and similarity summation from the previous architecture are retained to produce the final similarity value, as illustrated in Figure \ref{fig:embm} (A). By incorporating these modifications, the approach is expected to enhance both precision and recall. More correspondences can be identified by aligning synonyms, while precision increases as more homonym false positives are filtered out. Moreover, this modification increases the adaptability and flexibility of the proposed approach framework, as new state-of-the-art embedding models can be applied to it without changes in its architecture.

\subsubsection{Embeddings of SPARQL query (ESQ)}
\label{sec:query-emb}

Another improvement in the architecture builds upon the previous one and focuses on the aggregation of SPARQL query embeddings in the similarity comparison in Step 7 (Figure \ref{fig:narch}). In the previous configuration, the embeddings from the labels are individually compared in a cross-product manner, and the final similarity is computed as the summation of each similarity after filtering. While the initial modification enhanced semantic comparison by employing individual embeddings for each label, aggregating the SPARQL query embeddings before comparison can offer a more contextual representation of the SPARQL query and enhance the quality of the correspondences identified. The proposed modification involves aggregating embeddings associated with each label extracted from the SPARQL query by averaging them once before all similarity comparisons with labels retrieved from target subgraphs. Once the aggregated embedding for the SPARQL query is obtained, it is compared with the embeddings of labels retrieved from target subgraphs using cosine similarity, similar to the previous modification. However, instead of comparing individual embeddings, the aggregated embedding is compared directly with the embeddings of labels in the subgraphs, and the threshold filter is also applied. This process is illustrated in Figure \ref{fig:embm} (B). It offers advantages over the previous one since aggregating embeddings can capture a richer semantic context of the SPARQL query. Comparing the embeddings individually may lack essential information to identify some correspondences, which can be overcome by aggregating the embeddings beforehand.

\subsubsection{Subgraph embeddings (SE)}

Inspired by graph neural network embeddings \cite{DBLP:journals/tnn/WuPCLZY21} that aggregate node embeddings to represent graphs and subgraphs, this modification aims to capture the collective semantic information embedded within subgraphs. It entails aggregating embeddings associated with labels extracted from subgraphs, considering both unary and binary queries. In the baseline approach, unary and binary queries generate different types of subgraphs, and the embeddings of these subgraphs vary accordingly. For unary SPARQL queries, the subgraph consists of triples composed of subject, predicate, and object. In addition, subjects and objects are associated with \textit{subjectType} and \textit{objectType}, respectively, which correspond to the most similar type of the corresponding entity determined by employing embedding similarity in the entity labels. 

\begin{figure}
    \centering
    \includegraphics[width=\textwidth]{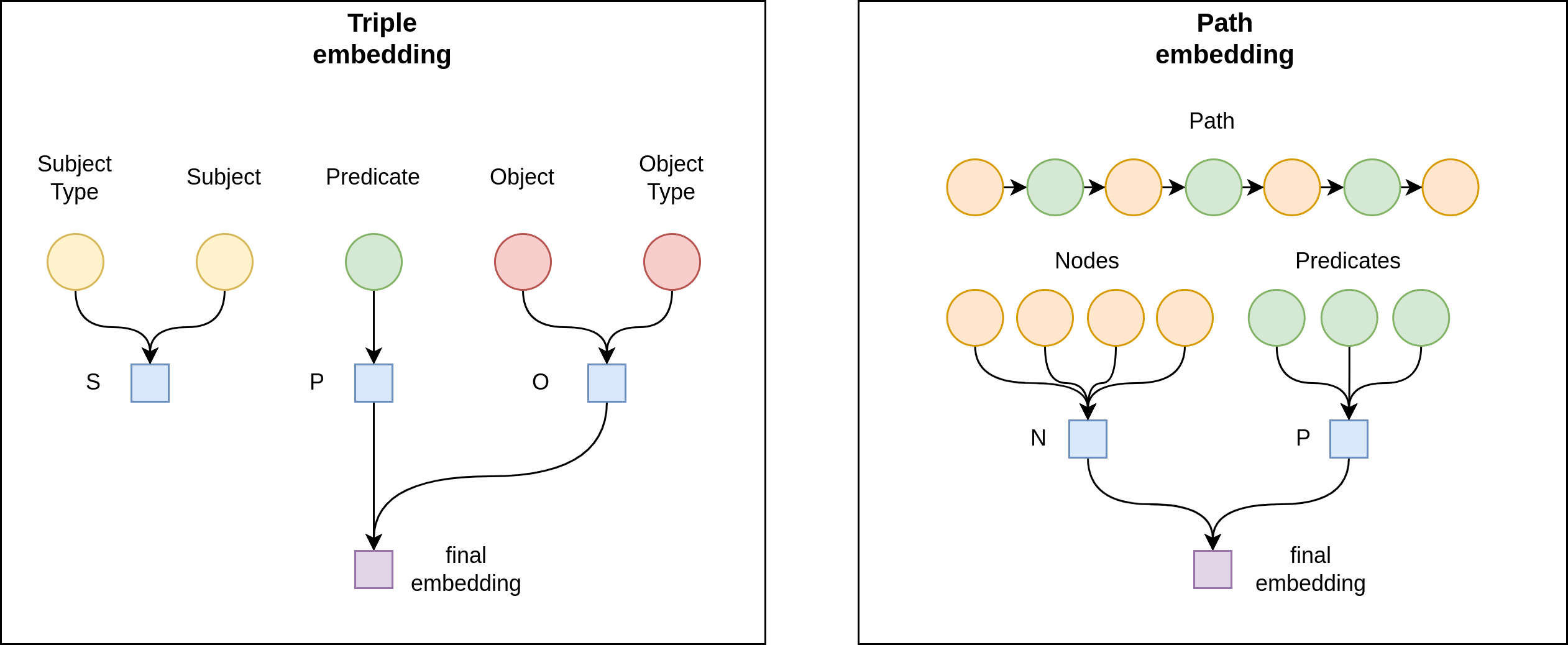}
    \caption{Example of a subject-type triple embedding, where the predicate embedding P and object embedding O form the final embedding. For predicate-type triples, the embeddings S and O are combined and for object-type triples, the embeddings P and O are combined. In binary queries the subgraphs are paths. The embedding of the nodes and predicated are aggregated independently and then the resulting embedding is aggregated to produce the final embedding representing the path.}
    \label{fig:triplex}
\end{figure}

{Each triple can have a specific type based on the location of the instance in subject, predicate, and object. The final embedding for each type uses the embeddings of the complement information to produce the final embedding. To create the embedding for the subject and object, the labels of the subject entity and \textit{subjectType} (or \textit{objectType} for object) are embedded and averaged, similar to the SPARQL query embedding method and averaged again with the predicate as shown in Figure \ref{fig:triplex} and Figure \ref{fig:embm} (C).} 
For binary SPARQL queries, the subgraph denotes a path, which is a sequence of entities linked by properties. The final embedding is constructed by initially aggregating all embeddings of nodes, followed by aggregating all embeddings of properties. Finally, the aggregated embedding of the nodes is combined with the aggregated embedding of properties to create a representation of the entire path, as illustrated in Figure \ref{fig:triplex}. 

Once the aggregated embeddings for both the SPARQL query and subgraph labels are obtained, they are compared using cosine similarity, similar to the previous modifications. However, rather than comparing each embedding individually with each other, the aggregated embeddings of the subgraph are directly compared with the aggregated SPARQL query embedding. It is expected to produce a more contextualized representation of the subgraphs. Since only a single embedding is compared with the SPARQL query embedding, a reduction in runtime is also expected. However, as the similarity threshold is applied in only one comparison, it has the potential to reject the entire subgraph if the similarity value falls below the threshold.

\section{Experiments}
\label{sec:experiments}

\subsubsection{Dataset} The dataset used in this experiment consists of the populated version of the OAEI Conference benchmark\footnote{\url{http://oaei.ontologymatching.org/2024/complex/index.html}}. It comprises 5 ontologies and 100 manually generated SPARQL queries. The evaluation is conducted in pairs, yielding a total of 20 evaluation pairs. For each pair, the SPARQL queries were taken from the source ontology, and thus, the order of the ontologies in the pair counts. This dataset has been chosen as it is equipped by CQAs and is already used for running several systems, which allows for comparisons. 

\subsubsection{Models}

The LLMs were selected from the Massive Text Embedding Benchmark (MTEB) \cite{DBLP:conf/eacl/MuennighoffTMR23}, which serves as a benchmark for evaluating various language models across diverse text embedding similarity tasks such as clustering, retrieval, and classification. The proposed approach is tested with the 11 best-performing models on average. Not all models were selected strictly based on their leaderboard order. Some models were excluded due to requiring a paid API subscription where the free subscription proved insufficient for our experimental needs, while others were omitted because they exceeded the capacity of the GPUs available to us at the time. The leaderboard of models was accessed through \url{https://huggingface.co/spaces/mteb/leaderboard} on 13/03/2024. These models were loaded using the Transformers library \cite{DBLP:conf/emnlp/WolfDSCDMCRLFDS20} from Hugging Face. In addition to the selected LLMs, smaller embedding models are included for comparison purposes: BERT-INT \cite{DBLP:conf/ijcai/Tang0C00L20} in the entity link step; and word embeddings such as GloVe \cite{DBLP:conf/emnlp/PenningtonSM14}, Word2Vec \cite{DBLP:journals/corr/abs-1301-3781}, and FastText \cite{DBLP:journals/tacl/BojanowskiGJM17} in other steps, resulting in a total of 15 models. The selected models are presented in Table \ref{tab:models}.

\begin{table}[!ht]
    \centering
    \begin{tabular}{|l|r|}
        \hline
        Embedding Model & Dimension \\
        \hline
        BERT-Int & 768 \\
        BGE-base-en-v1.5 (BAAI) & 768 \\
        BGE-large-en-v1.5 (BAAI) & 1024 \\   
        E5-mistral-7b-instruct (intfloat) & 4096  \\
        Echo-mistral-7b-instruct-lasttoken (jspringer) & 4096  \\
        Ember-v1 (llmrails) & 1024   \\
        Fasttext & 300   \\ 
        Glove & 300  \\           
        GritLM-7B (GritLM) & 4096   \\
        GTE-large (thenlper) & 1024   \\      
        Mxbai-embed-large-v1 (mixedbread-ai) & 1024  \\
        SFR-Embedding-Mistral (Salesforce) & 4096   \\      
        Stella-base-en-v2 (infgrad) & 768   \\
        UAE-Large-V1 (WhereIsAI) & 1024    \\      
        Word2vec & 300 \\
        \hline
    \end{tabular}
    \caption{Language models selected for experimentation.}
    \label{tab:models}
\end{table}

\subsubsection{Metrics}

The evaluation metrics used here are the ones adopted in the OAEI campaigns for the Populated Conference dataset. These metrics are based on the comparison of instance sets. The generated alignment by the systems is used to rewrite a set of reference source SPARQL queries whose results (set of instances) are compared to the ones returned by the corresponding target reference SPARQL query. This comparison shows the overall $coverage$ of the generated alignment concerning the knowledge needs and the best-rewritten query\footnote{The description of rewriting systems is out of the scope of this paper. For details, the reader can refer to {\cite{DBLP:journals/semweb/ThieblinHT21}}.}. A balancing strategy calculates the intrinsic alignment $precision$ based on common instances. Given an alignment $A_{eval}$ to be evaluated, a set of SPARQL query reference pairs $query_{pairs}$ (composed of source $query_s$ and target $query_t$), $kb_s$ the source knowledge base, $kb_t$ a target knowledge base, and $f$ an instance set ($I$) comparison function:

\vspace{-0.5cm}
\begin{multline}
\label{eq:coverage}
cov(A_{eval},query_{pairs},kb_s,kb_t,f) = 
\underset{\langle query_s,query_t \rangle\in query_{pairs}}{\text{average}} f( I_{query_t}^{kb_s}, I_{best q_t}^{kb_t}) 
\end{multline}

Different functions $f$ can be used to compare the similarity of instance sets (overlap, precision-oriented, recall-oriented, etc.). Here, $coverage$ is based on the $queryFmeasure$ (also used for selecting the best-rewritten query). This is motivated by the fact that it better balances precision and recall. Given a reference instance set $I_{ref}$ and an evaluated instance set $I_{eval}$:

\begin{equation}
\label{eq:prec_rec}
QP=\frac{|I_{eval} \cap I_{ref}|}{|I_{eval}|} 
\qquad
QR=\frac{|I_{eval} \cap I_{ref}|}{|I_{ref}|}
\end{equation}
\begin{equation}
\label{eq:query_fmeasure}
queryFmeasure(I_{ref},I_{eval})= 2 \times \frac{QR\times QP}{QR+QP}
\end{equation}
\begin{multline}
\label{eq:bestcq}
best q_t= \underset{ q_t\in rewrite(query_s,A_{eval},kb_s)}{\text{argmax}} queryFmeasure( I_{query_t}^{kb_t},I_{q_t}^{kb_t})
\end{multline}

Balancing $coverage$, $precision$ is based on classical (i.e., scoring 1 for same instance sets or 0 otherwise) or non-disjoint functions $f$: 
\begin{equation}
\label{eq:int_prec}
precision(A_{eval},kb_s,kb_t,f) = \underset{\langle e_s, e_t \rangle\in A_{eval}}{\text{average}} f(I_{e_1}^{kb_s}, I_{e_2}^{kb_t} )
\end{equation}

\subsubsection{Parameters}

The proposed approach takes several inputs, including the source ontology path, target ontology path, output folder, link for embedding files, similarity threshold, architecture setting, and link similarity (if embeddings are used in the link step). There are 20 possible pairs for both source and target ontologies. In addition, 6 possible architecture settings can be combined with 2 possible link types.  With the selected models, there are 15 choices, leading to a total of 3600 possible combinations. To manage the running time and explore parameter impacts effectively, 1800 combinations are considered without embeddings in the link step initially. Then, the best models and architecture are used in the link step.
For embeddings, the similarity range spans from 0.5 to 1.0, increasing by 0.1 increments. Regarding link types, thresholds of 0.8, 0.85, and 0.9 were utilized. The final results were computed by running the evaluator on each alignment file, followed by selecting the pair with the best F-measure per threshold. Finally, all values were averaged by each model.

The embeddings are computed once for all datasets using a GPU with 24GB VRAM and are cached on disk. During the matching process, these precomputed embeddings are loaded from the cache, and no GPU is used. For reading the embeddings, the matcher reads the text file and stores each embedding in a key-value store. Here, the key denotes the label used to generate the embedding, while the corresponding value is the embedding generated from the model. To manage the embeddings, perform aggregations, and compute operations such as cosine similarity, the library Deeplearning4j\footnote{https://deeplearning4j.konduit.ai/ access at 02/04/2024.} was used. These operations were directly computed by the matcher when required.

\section{Results and Discussion}
\label{sec:results}

\begin{figure}[ht!]
    \centering
    \includegraphics[width=\textwidth]{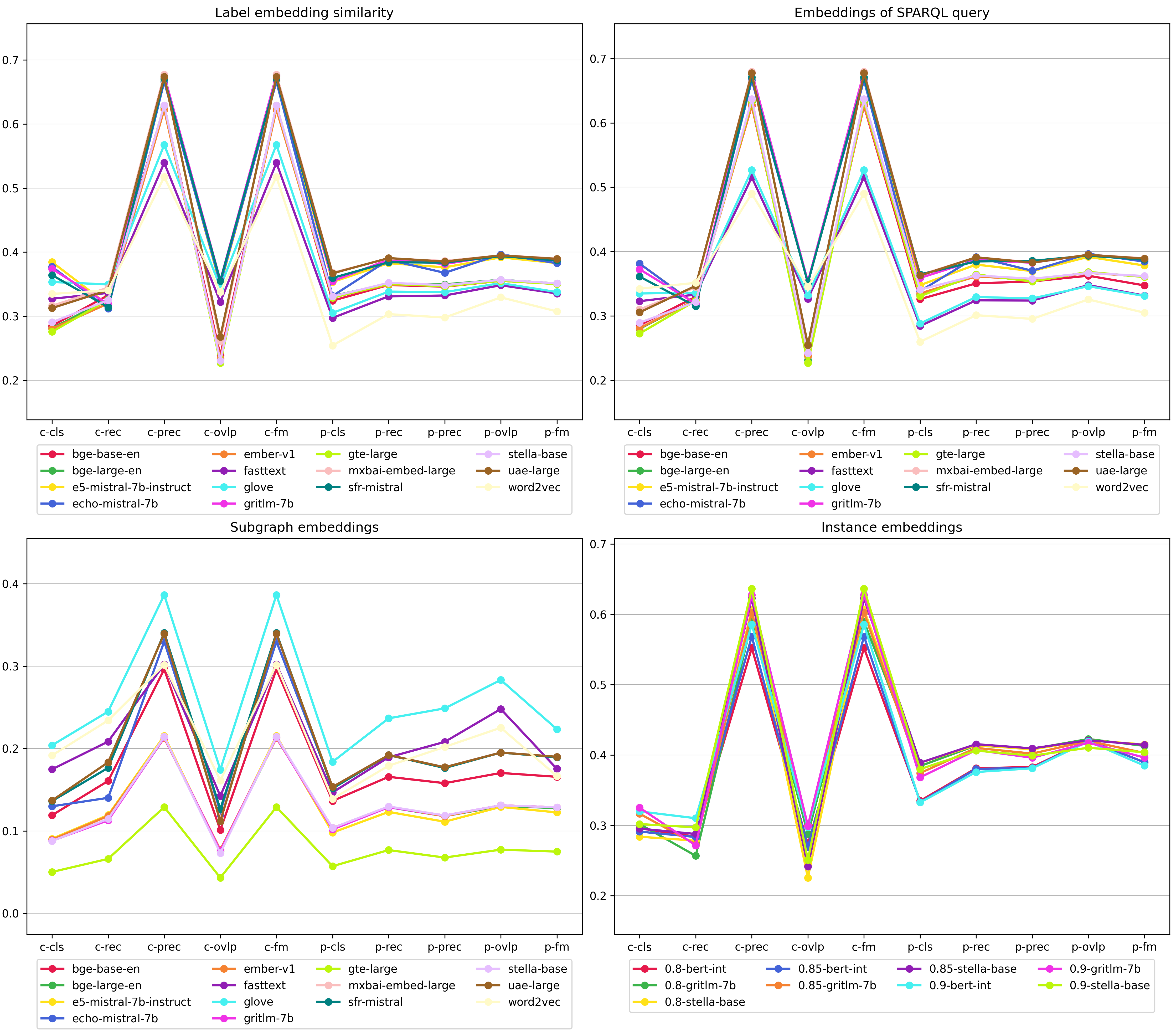}
    \caption{Performance of the models when used with each architecture setting.}
    \label{fig:lmch}
    \vspace{-0.5cm}
\end{figure}

{The analysis of all models in all settings is presented in Figure \ref{fig:lmch}. The highest results are observed from LES and ESQ. In both settings, increasing the size of the models also enhances the system's performance. The setting with the poorest results is SE. In this setting, the embeddings with some LLMs exhibit reduced performance, while GloVe embeddings demonstrate less degradation. Additionally, the setting IE improves the precision of the matcher in all tested models while reducing the query-based metrics in some models.}

The results of the best model in all architecture settings are presented in Table \ref{tab:brs}. The setting achieving the highest query-oriented classical and query-oriented overlap is LES with embeddings from the LLM GritLM-7B, reaching 0.37 compared to the baseline's 0.35 and 0.36 compared to the baseline's 0.35, respectively. In query-oriented recall, the baseline still achieves the best results with 0.36 compared to the second-place ESQ (ignore case) with SFR-Mistral reaching 0.32. For query-oriented precision and query-oriented F-measure, the ESQ setting with the LLM GritLM-7B reached the best results with 0.68 in precision and 0.68 in F-measure compared to the baseline's 0.47 in both metrics. In precision-oriented evaluation, the setting in the IE with Stella-Base achieved the highest results in all metrics.

\begin{table}[!ht]
    \centering
    \begin{tabular}{|l|c|c|c|c|c|c|c|c|c|c|c|}
    \hline
    & \multicolumn{5}{c}{query oriented} & \multicolumn{5}{|c|}{precision oriented}\\
    \hline
    model & cls & rec. & prec. & ovlp & f-m. & cls & rec. & prec. & ovlp & f-m.\\
    \hline
    base (levenshtein) & 0.35 & \textbf{0.36} & 0.47 & 0.35 & 0.47 & 0.21 & 0.26 & 0.26 & 0.28 & 0.26 \\
    GritLM-7B (LES) & \textbf{0.37} & 0.32 & 0.68 & \textbf{0.36} & 0.68 & 0.36 & 0.39 & 0.38 & 0.40 & 0.39 \\
    sfr-mistral (i-LES) & 0.37 & 0.32 & 0.67 & 0.35 & 0.67 & 0.36 & 0.39 & 0.38 & 0.39 & 0.39 \\
    GritLM-7B (ESQ) & 0.37 & 0.32 & \textbf{0.68} & 0.35 & \textbf{0.68} & 0.36 & 0.39 & 0.38 & 0.40 & 0.39 \\
    sfr-mistral (i-ESQ) & 0.37 & 0.32 & 0.67 & 0.35 & 0.67 & 0.36 & 0.39 & 0.38 & 0.39 & 0.39 \\
    glove (SE) & 0.20 & 0.25 & 0.39 & 0.17 & 0.39 & 0.18 & 0.24 & 0.25 & 0.28 & 0.22 \\
    glove (i-SE) & 0.21 & 0.25 & 0.40 & 0.18 & 0.40 & 0.18 & 0.23 & 0.25 & 0.28 & 0.22 \\
    stella-base (ESQ+IE 0.9) & 0.30 & 0.30 & 0.64 & 0.25 & 0.64 & 0.38 & 0.41 & 0.40 & 0.41 & 0.40 \\
    stella-base (ESQ+IE 0.85) & 0.30 & 0.29 & 0.62 & 0.24 & 0.62 & \textbf{0.39} & \textbf{0.42} & \textbf{0.41} & \textbf{0.42} & \textbf{0.41} \\
    GritLM-7B (ESQ+IE 0.9) & 0.33 & 0.27 & 0.63 & 0.30 & 0.63 & 0.37 & 0.41 & 0.40 & 0.42 & 0.40 \\
    \hline
    \end{tabular}
    \caption{Results for best models in each setting. \textit{i} in the setting name refers to ignore case version. The values near IE are the threshold applied in the link step.}
         \label{tab:brs}
\end{table}

As depicted in Table \ref{tab:brs}, LES and ESQ exhibit the highest results when LLMs are applied, both individually and on average with the other models. These settings involve fewer embedding aggregations. It is also evident that as the number of aggregations increases, as in the SE configurations, the results of all models deteriorate. One of the reasons for this degradation is that combining embeddings without a weight transformation, such as Graph Neural Networks, can cause the embeddings to lose semantic information and increase noise. In all settings, increasing the size of the models leads to improved results, although among the LLMs, the results were similar. Another observation is that using the setting of IE improves the precision-oriented metrics of all the models. Furthermore, while query-oriented metrics increased in some models, they decreased in others. A comparison with state-of-the-art complex matches was also performed and the results of the new matcher are summarized in Table \ref{tab:ots}. Employing LLM embeddings and the new architecture settings, the proposed matcher outperforms in both precision and F-measure compared to the other approaches.

\begin{table}[!ht]
    \centering
    \begin{tabular}{|l|c|c|}
        \hline
        Matcher & Prec. & Coverage \\ 
        \hline
        Matcha-DL & - & - \\
        AMLC & 0.230 & 0.260 \\
        CANARD & 0.212 & 0.471 \\
        Our (Stella-base IE 0.85) & \textbf{0.389} & 0.623 \\
        Our (GritLM-7B ESQ) & 0.359 & \textbf{0.679} \\
        \hline
    \end{tabular}
    \caption{Comparison of the proposed approach with other matchers in the state of the art. Prec. in the table refers to (classical - not disjoint) precision and Coverage to (classical - query F-measure) coverage. Results for Matcha-DL are not present since it didn't produce results in the Complex Track.}
    \label{tab:ots}
    \vspace{-0.75cm}
\end{table}

\subsubsection{Impact of each setting}

Another experiment conducted is the impact of each setting when using the same LLM to see how the aggregation of embeddings and also how the use of IE impacts the matcher performance. In this experiment, the LLM GritLM was considered and evaluated with all. The results of this evaluation are in Table \ref{tab:prog}. The values from the  LES and Embeddings of the SPARQL query are quite similar. The SE reduces the results in both precision and query-oriented. In the IE the results of the precision-oriented increase while the query-oriented is reduced compared to the best approaches. Also, the values in the query-oriented and precision-oriented IE settings follow an inverse relation with the change of threshold, when the threshold increases the query-oriented values increase while the precision-oriented values decrease. The last analysis is regarding the runtime of each setting. The experiments were conducted using the same LLM embeddings in the same ontology pair between cmt and conference varying only in the architecture setting. All the experiments were run on CPU without GPU acceleration. The fastest setting is the baseline approach with 29 seconds. The ESQ, SE, and IE take around 2 minutes. Among these, the ones with more aggregations have reduced runtime. The setting with the highest runtime is the IE with 2 hours and 11 minutes. This high runtime is due to the need to get the similarity of all instances in the target knowledge graph to get the equivalent instances in the link step.

\begin{table}[!ht]
\centering
\begin{tabular}{|l|l|c|c|c|c|c|c|c|c|c|c|c|}
\hline
\multicolumn{2}{|c}{} & \multicolumn{5}{|c}{query oriented} & \multicolumn{5}{|c|}{precision oriented}\\
\hline
model & config. & class & rec. & prec. & overlap & f-m. & class & rec. & prec. & overlap & f-m.\\
\hline
GritLM-7B & LES & 0.37 & 0.32 & 0.68 & 0.36 & 0.68 & 0.36 & 0.39 & 0.38 & 0.40 & 0.39 \\
GritLM-7B & ESQ & 0.37 & 0.32 & 0.68 & 0.35 & 0.68 & 0.36 & 0.39 & 0.38 & 0.40 & 0.39 \\
GritLM-7B & SE & 0.09 & 0.11 & 0.21 & 0.08 & 0.21 & 0.10 & 0.13 & 0.12 & 0.13 & 0.13 \\
GritLM-7B & IE (th 0.8) & 0.30 & 0.26 & 0.59 & 0.29 & 0.59 & 0.38 & 0.41 & 0.41 & 0.42 & 0.41 \\
GritLM-7B & IE (th 0.85) & 0.32 & 0.27 & 0.60 & 0.30 & 0.60 & 0.37 & 0.41 & 0.40 & 0.42 & 0.40 \\
GritLM-7B & IE (th 0.9) & 0.33 & 0.27 & 0.63 & 0.30 & 0.63 & 0.37 & 0.41 & 0.40 & 0.42 & 0.40 \\
\hline
\end{tabular}
\caption{Progressive evaluation of GritLM in all architecture. In the config. column, LES refers to Label embedding similarity, ESQ to Embeddings of SPARQL query, SE to Subgraph embeddings, and IE Instance embeddings.}
\label{tab:prog}
\vspace{-0.5cm}
\end{table}

\section{Related Work}
\label{sec:related_work}

Previous work in ontology matching has explored the use of embeddings and language models. The review in \cite{DBLP:conf/semweb/SousaLT22} categorizes ontology matching approaches according to how information is incorporated into embeddings. More recently, LLMs have also been exploited \cite{10.1145/3587259.3627571,DBLP:conf/adbis/PeetersB23,DBLP:journals/pvldb/ZeakisPSK23,DBLP:journals/vldb/LiLSDT23}, where LLMs extract matching candidates through prompt engineering techniques. However, very few approaches have still addressed the issue of complex matching in ontologies. Matcha-DL \cite{DBLP:conf/om2/FariaSCFBP23} is an ontology matching approach that addresses both simple and complex ontology matching. It employs multiple metrics to identify alignments between ontologies, including lexical and structural metrics, background knowledge, and pattern-based complex matching. It also incorporates an LLM module to generate embeddings for entity labels, enhancing similarity computation, along with a translation module based on deep learning models. In this approach, a neural network is further employed to combine all similarity metrics, producing the final similarity score, which can be refined through training. Besides its potential, the matcher is still not evaluated in the complex matching task, and is difficult to compare its performance with other systems. 

In \cite{DBLP:conf/i-semantics/DhouibFT19,DBLP:journals/jodsn/DhouibFT21}, a complex matching approach is proposed, which applies word vector embeddings to encode entity labels that are subsequently refined with a radius measure. The initial step in this approach involves generating the embedding representation of entities. This representation is created using fastText \cite{DBLP:journals/tacl/BojanowskiGJM17} to embed each word within the textual information of the entity. Given that the textual information of an entity can comprise multiple words, an aggregation step is performed to merge the word embeddings into a single embedding for each entity. Following this step, a contextualized representation is generated for each entity by treating each entity as a cluster root. Then, the embeddings of the entities within its hierarchy are aggregated to generate the cluster embedding. Finally, a measure of the cluster's radius is employed to refine the relationship between the entities. For instance, if an entity is close to the cluster center, it indicates a broader and more general relation exists between them, as the radius encapsulates the concepts of all entities within the cluster. But in this work, the embedding models used are static and as shown in the results, they have worse performance compared to LLMs and transformer-based models. 

In \cite{DBLP:conf/dis/AkremiAZ22}, a fuzzy ontology embedding is applied to produce embeddings for complex matching. It employs a fuzzification strategy to generate a fuzzy representation of the ontology, which is subsequently encoded using a similar strategy to RDF2Vec. A series of random walks generate a document of sentences that are then input into Word2Vec to generate the final word embeddings. Then, embeddings are used to calculate the similarity between entities. Later, in \cite{DBLP:conf/inista/AkremiAZ23}, the above approach is improved to project the embeddings using a hypothesis graph \cite{DBLP:journals/biomedsem/AgibetovJOSBGCO18} that is subsequently encoded using traversal walking methods such as in RDF2Vec \cite{DBLP:conf/semweb/RistoskiP16}, and node2vec \cite{DBLP:conf/kdd/GroverL16} to generate sentences encoded with Word2Vec and fastText \cite{DBLP:journals/tacl/BojanowskiGJM17}. Besides the proposition of a complex matcher, the models based on random graph traversal are known to predict the relatedness instead of similarity \cite{DBLP:conf/esws/PortischCSKHP22}, as needed in the matching task reducing its performance.

In few works \cite{DBLP:conf/i-semantics/DhouibFT19,DBLP:journals/jodsn/DhouibFT21,DBLP:conf/dis/AkremiAZ22} embeddings are used in complex matching. However, the embeddings applied are still word embeddings like Glove that have less performance than the Transformer model embeddings, as observed in the experiments with these models in this work. Also, these works do not apply recent models like LLM which produces better representations of natural language. Also, in Matcha-DL \cite{DBLP:conf/om2/FariaSCFBP23} LLMs are applied in the matcher pipeline and the approach can be trained using data in a supervised manner. However, for the task, still few datasets with reference alignments for training are available and this type of data can be difficult to find and produce. For that reason, pre-trained models and unsupervised techniques are still preferred for this task as applied in this work. More recently, \cite{DBLP:journals/corr/abs-2404-10329} has proposed to use LLM to produce complex alignments where ChatGPT-4 is used to produce alignments between the ontologies GMO and GBO. The ontologies are modularized and the LLM is prompted to find the common modules between the two ontologies. However, this approach is effective while the complex entity modules are present and the majority of ontologies aren't modularized. The alignment response of the LLM contains natural language text that difficult automatic evaluation. 

\section{Conclusions and Future Work}
\label{sec:conclusions}

This paper has proposed to integrate LLMs into an approach for generating expressive correspondences based on alignment need and ABox-based relation discovery. The proposed approach has achieved superior results in nearly every metric compared to the baseline without embeddings, as well as improvement over other state-of-the-art systems in this task. The approach does not require any training or reference alignments. Also, the models used to generate the embeddings are not fine-tuned improving the approach capacity of generalization.
The approach can be extended in several directions. The guidance provided by user needs is both a strength and a limitation of the approach (it facilitates generalization across a limited number of instances but requires users' ability to express their needs as SPARQL queries). 
The first direction for extension involves devising a purely T-Box strategy. Second, the problem could be subdivided into sub-tasks through ontology partitioning, given the inherently vast search space of the task. Third, exploring improved aggregation techniques for subgraphs may yield superior results. Finally, fine-tuning LLMs and delving deeper into the prompts guiding the creation of entity embeddings can also be addressed.

\bibliographystyle{splncs04}
\bibliography{ekaw}

\end{document}